\documentclass{article}

\usepackage{arxiv}

\usepackage[utf8]{inputenc} 
\usepackage[T1]{fontenc}    
\usepackage{hyperref}       
\usepackage{url}            
\usepackage{booktabs}       
\usepackage{amsfonts}       
\usepackage{nicefrac}       
\usepackage{microtype}      
\usepackage{lipsum}
\usepackage{graphicx}
\graphicspath{ {./images/} }

\title{Chili Pepper Disease Diagnosis via Image Reconstruction Using GrabCut and Generative Adversarial Serial Autoencoder}

\author{
 Jongwook Si \\
  Dept. Computer AI Convergence Engineering\\
  Kumoh National Institute of Technology\\
  Gumi, KOREA 39177 \\
  \texttt{jwsi425@kumoh.ac.kr} \\
   \And
 Sungyoung Kim \\
  Dept. Computer Engineering\\
  Kumoh National Institute of Technology\\
  Gumi, KOREA 39177 \\
  \texttt{sykim@kumoh.ac.kr} \\
  \And
}

\begin{document}
\maketitle
\begin{abstract}
With the recent development of smart farms, researchers are very interested in such fields. In particular, the field of disease diagnosis is the most important factor. Disease diagnosis belongs to the field of anomaly detection and aims to distinguish whether plants or fruits are normal or abnormal. The problem can be solved by binary or multi-classification based on CNN, but it can also be solved by image reconstruction. However, due to the limitation of the performance of image generation, SOTA's methods propose a score calculation method using a latent vector error. In this paper, we propose a network that focuses on chili peppers and proceeds with background removal through Grabcut. It shows high performance through image-based score calculation method. Due to the difficulty of reconstructing the input image, the difference between the input and output images is large. However, the serial autoencoder proposed in this paper uses the difference between the two fake images except for the actual input as a score. We propose a method of generating meaningful images using the GAN structure and classifying three results simultaneously by one discriminator. The proposed method showed higher performance than previous researches, and image-based scores showed the best performance.
\end{abstract}


\section{Introduction}
Smart Farm refers to an intelligent farm that can automatically manage the growing environment of crops and livestock by applying convergent technologies such as unmanned automation, artificial intelligence, and big data to greenhouses and livestock. It is a technology that can be properly prescribed anytime, anywhere through the judgment of the environment or biometric information of production facilities, and can maximize productivity through minimal labor and energy. These smart farms contribute to production and distribution using environmental sensors, crop sensors, and soil sensors. In particular, in the case of Korea, there is a risk of the sustainability of related industries due to the aging of the population engaged in the agricultural and livestock industries and the decrease in the inflow of young people. In particular, the overall production population is decreasing due to the decline in the fertility rate, which has emerged as a social problem, and in particular, the elderly population of farmers (age 65 or older) is continuously in-creasing every year. Therefore, smart farms are recognized as a key means to strengthen agricultural competitiveness and secure various age groups of farmers through youth in-flow. In particular, artificial intelligence is solving many challenges facing agriculture and uses technologies such as machine learning, computer vision, and predictive analysis. The most productive aspect of smart farms is the crop. In order to manage crops in a smart farm environment, it is necessary to continuously create an environment that is most appropriate for the crop, and the diseased crops should be excluded from the production. In this paper, we want to pay attention to chili peppers among various crops, and the purpose of this paper is to check the condition of the crops using computer vision technology to classify whether they are normal or diseased.
There are two main methods of determining the state of crops using computer vision technology. The most frequently used method is to determine the input image after converting the result for the last layer through several convolution layers to a value between 0 and 1 through activation function. Also, based on the autoencoder structure, image reconstruction is performed with an image that follows a normal data distribution using only normal data, and the state of the crop can be determined using an error in input/output results. However, the image reconstruction method is relatively low in accuracy because it is difficult to reconstruct accurately. Therefore, research was conducted to improve performance by grafting the autoencoder structure to GAN. Due to the limitation in accuracy, although performance improvement has been achieved more than before, recent research-es have made significant performance improvement using latent vector between encoder and decoder. In this paper, we present a novel framework that overcomes the limitations of performance improvement in image domain rather than the latent vector and has high-er accuracy.

The contribution of this paper is as follow.

•	We present a new GAN-based framework for detecting crop anomalies using errors between image domains in image reconstruction.\\
•	By connecting two auto encoders in series, a new method of diagnosing normal and diseased using two reconstructed images is presented, and the performance is higher than that of previous researches.\\
•	The background of the chili pepper image may be removed through a grabcut algorithm in the image processing field to reduce a result of misclassification.

\section{Related Works}
\label{sec:headings}
After GAN[1] was first presented, many researchers began to develop and apply it. This structure, in which the generator and the discriminator are opposed to each other to improve performance, was mainly used in the image generation area. Although anomaly detection was still common to classify via image processing area or CNN, researchers began to apply GAN to anomaly detection, deviating from existing ideas.

\subsection{Anomaly Detection via Image Reconstruction}
AnoGAN[2] is an unsupervised learning that uses only normal image data as learning as a study first used for the purpose of detecting outliers. This study with a GAN-based structure adds Residual Loss to allow generators to train manifolds of normal data and to reduce the difference between inputs and outputs. Discrimination Loss includes a Featuring Matching process that identifies the probability distribution of input data, identifies real and fake, and proceeds with learning to follow the distribution of normal data. It was noted as the first idea based on GAN about anomaly detection, many GAN-based anomaly detection researches have been actively conducted. By improving this AnoGAN[2], f-AnoGAN[3] of 2-step method in which encoder training is added after GAN learning has emerged. f-AnoGAN[3] conducted encoder training for latent space mapping of data to obtain the result of feature extraction, and generated the result of making it the input of the generator again to improve speed and performance. To overcome the short-comings of autoencoder, which is difficult to find fine results due to reconstruction of ab-normal areas, it was solved by searching for and generating the most relevant items based on the memory of normal data using memAE[4]. SALAD[5] proposed a novel anomaly detection framework with image reconstruction that considers both image and latent spaces. By adding a loss function using SSIM and a loss function that constrains the center, only useful information can be reconstructed from normal data. SSM[6] is a framework that adds random masking and restoration to the autoencoder structure, which proposes a method to improve image reconstruction results. This research improves the learning of inpainting and has the ability to locate abnormal areas through masks of various sizes. J. Liu et. al.[7] presented a GAN-based network for fiber defect detection. The proposed net-work is trained in multistage and has been confirmed to have high defect detection performance under various conditions. J. Si et. al.[8] proposed a method of determining defects through an image reconstruction method of a thermal image of a solar cell. This re-search conducted an evaluation using only some pixels of the subtraction image that best represented the characteristics, and showed higher performance than the patch method. Another study by the same author [9] applies Fourier transform to the resulting image to solve the problem that the image reconstruction results of the AutoEncoder structure differ greatly from the input. The applied results only preserve features with large differences and improve performance against anomaly detection.

\subsection{Diseased Plant Diagnosis}
DAGAN[10] proposed research in the field of anomaly detection that solving the problem of imbalance between samples. DAGAN[10] introduced an autoencoder structure not only to the generator but also to the discriminator, improving the stability of learning and the performance of image reconstruction. DoubleGAN[11] proposed a GAN structure to detect the disease of plants. This research consists of two steps. Step 1 uses both normal and diseased plants, and based on WGAN [12], data augmentation is carried out using a pretrained model. Subsequently, in Step 2, the number of diseased leaves data are expand-ed by increasing the size of the image by 16 times using SRGAN[13]. It uses several ex-tended GANs to improve performance. Punam Bedi et. al.[14] proposed a hybrid model for detecting plant diseases by combining CAE and CNN. It creates a network that can reconstruct images through the CAE structure and simultaneously focuses on the compressed domain representation, the output of the encoder. This domain is a latent vector, which is used to classify normal or diseased by CNN. This study achieved high performance with only 9,914 parameter training. Sharath D. M. et.al.[15] proposed a framework for detecting plant diseases using data-preprocessing and CNN for image processing. They use with the grabcut segmentation proposed by Yubing Li et.al.[16] for background removal, which improved Grabcut[17]. And use morphology error to eliminate noise. This preprocessed data is provided to the user through a mobile application with nine labels through a CNN classifier. A. Abbas et.al.[18] introduces how to generate synthesized images using C-GAN[19] to detect diseases in tomatoes and differentiate 10 diseases through transfer learning in DenseNet121[20]. This method shows very high classification accuracy and the effect of data augmentation on classification performance improvement. R. Katafuchi et. al.[21] proposed a method for detecting plant diseases based on the ability to restore colors. We use Pix2pix[22] to print the results of the original restoration and present a new Anomaly Score based on CIEDE2000[23] to improve performance. T. Tosawadi et. al.[24] proposed a method of detecting disease for rice plants. This research can improve the performance of disease recognition for small areas by selecting and classifying multiple sec-tions from diseased areas.

\section{Proposed Method}
\label{sec:headings}
The method proposed in this paper is based on GAN. We propose a novel method with serial autoencoders as generator and require training of image reconstruction methods. For disease diagnosis, we propose a scoring method using errors between images. In addition, the background is removed elements that interfere with chili pepper images, which uses Grabcut[17], an image processing technology.

\subsection{Preprocessing}
\label{sec:headings}
In order to reconstruct only useful information, the background is removed by leaving only chili peppers in the image. The technology used in this case is Grabcut[17]. Grabcut[17] is an algorithm based on the division of regions through the minimal cut algorithm used in the graph algorithm. Segmentation is possible by assuming the pixel of the image as the vertex of the graph and dividing it into foreground and background groups to find the optimal cut. The Grabcut[17]  process is shown in Figure 1.

\begin{figure}[htb!]
    \centering
    \includegraphics[width=15cm]{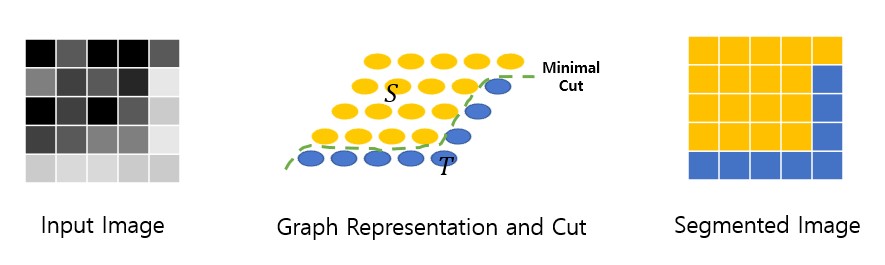}
    \caption{GrabCut Process}
\end{figure}

Strictly speaking, Grabcut[17] is a technique of extracting only objects corresponding to the foreground, and there is a point that feedback is possible through interaction with users. However, in this paper, since automatic disease diagnosis is required, we extract it by providing only areas with chili peppers, excluding interactions. However, since accurate results are not always produced, additional histogram-based work is carried out for precise foreground extraction. If the foreground does not appear for the original image, histogram equalization is performed on the entire image to change the distribution of pixels in the image.\\
The input of the deep learning network is the image of chili pepper region from which the background has been removed. A preprocessing is required to distribute the state of chili peppers the most. After removing noise and filling holes in the image through the morphology process, find the minimum rectangle rotated by the angle with the least background. An input image of 128x128 size is generated through perspective Transformation in this rectangle area, and the process is shown in Figure 2.

\begin{figure}[htb!]
    \centering
    \includegraphics[width=10cm]{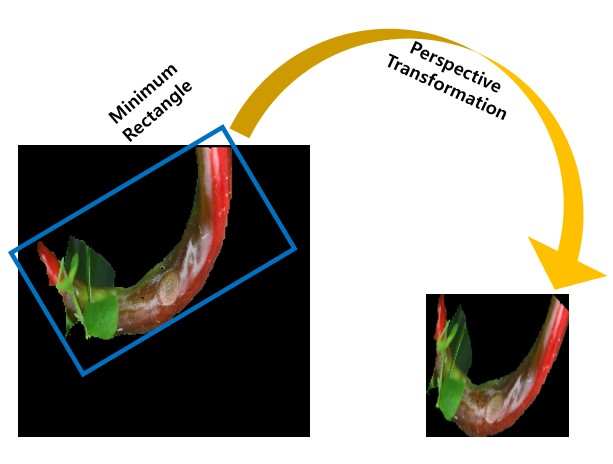}
    \caption{The process of perspective transformation by cropping only chili pepper area}
\end{figure}

\subsection{Approach Solution}
The method proposed in this paper uses Ganomaly[25] as Baseline, and shows improved results by overcoming the shortcomings of the previous image reconstruction re-searches anomaly detection of the image reconstruction method uses an autoencoder structure. This method is a method of compressing an input image into a latent vector and restoring the same. That is, if the network trains to compress and restore the normal input image, there is a assumption that the reconstructed image differs greatly when the abnormal image is input. However, even if the network preserves feature and implies low-level content, it is difficult to reduce the difference from the input image due to the limitations of the network. To solve this disadvantage, this paper shows a structure in which two auto encoders are connected in series (Figure 3).\\ 
Training about the network proposed in this paper takes place only on normal chili pepper images, and both normal and diseased chili pepper images are used in test. The generator generates two fake images and the results of which can be tricked by trying to produce similar results to the actual input images. The discriminator has the ability to simultaneously distinguish the input image and the two images generated by the generator. Therefore, the method using image reconstruction shows more improved results than previous researches using the score calculation method based on the latent vector.\\
The training datasets \(D_{trn}\) consists of only N normal chili pepper images that exist in large quantities. The test datasets \(D_{tst}\) consist of $M_1$ normal chili peppers and $M_2$ diseased chili peppers images. The labels on the normal datasets are all labeled zero because only normal images exist (i.e. \(D_{trn} = \{(x_1,y_1), \ldots, (x_N,y_N)\), where \(y \in \{0\}\)),and the test datasets are labeled zero for normal (\(x_{N_i}\)) and one for diseased (\(x_{D_i}\)) chili pepper images (i.e. \(D_{tst} = \{(x_{N_1},y_{N_1}), \ldots, (x_{N_{M_1}},y_{N_{M_1}}), (x_{D_1},y_{D_1}), \ldots, (x_{D_{M_2}},y_{D_{M_2}})\}\), where \(y \in \{0, 1\}\)). For training, we modify it to follow a standard normal distribution to approximate all the data like \(X \sim N(0,1)\), and we make the test data follow a normal distribution based on the mean and variance of the training datasets like \(D_{tst} \sim N(X_{tst} - \bar{X}, \sigma_X)\).

\begin{figure}[htb!]
    \centering
    \includegraphics[width=15cm]{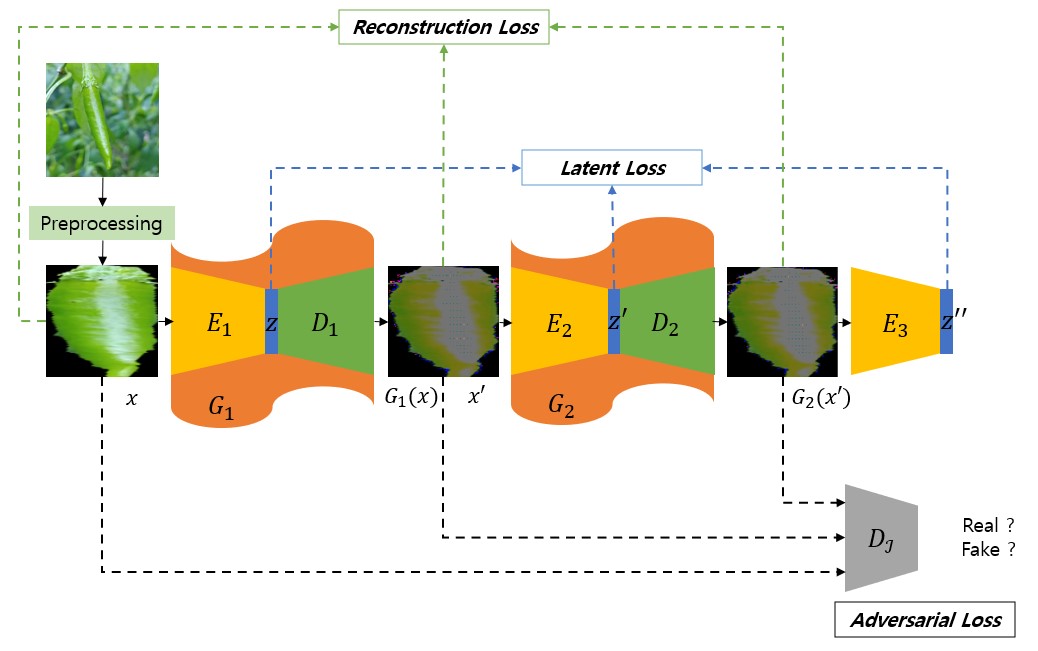}
    \caption{Overall training architecture}
\end{figure}

Since the generator (\(G_1, G_2\)) is composed of an autoencoder structure, it can be separated into an encoder and a decoder. The encoder (\(E_1, E_2, E_3\)) compresses the input image to generate a P-dimensional latent vector that represents it well, while the decoder (\(D_1, D_2\)) takes this latent vector (\(z, z', z''\)) as input to reconstruct an image closest to the original input of the encoder. \(z\), \(z'\), and \(z''\) represent the latent vectors that the encoder best expresses \(x\), \(x'\), and \(x''\), respectively. The difference between the latent vectors should also be reduced because the input image is compressed. Therefore, we have correlations such as \(z = E_1(x)\), \(z' = E_2(x') = E_2(G_1(x))\), and \(z'' = E_3(x'') = E_3(G_2(x'))\).

The first generator is defined as \(G_1\), and the second generator is defined as \(G_2\). \(G_1\) aims to produce real input images similar to the limits that the network can generate, while \(G_2\) is trained to produce the same results as \(G_1\). Consequently, we have correlations such as \(x' = G_1(x) = D_1(E_1(x))\) and \(x'' = G_2(x') = G_2(G_1(x)) = D_2(E_2(G_1(x)))\). By connecting two generators, \(G_1\) and \(G_2\), of this structure in series, it is possible to minimize errors between images.

The discriminator \(D_I\) serves to distinguish between real input images and the two results generated by the generators \(G_1\) and \(G_2\). From the standpoint of the discriminator \(D_I\), the real image should be determined as fake, and the image generated by the generators \(G_1\) and \(G_2\) should be identified as real. However, the generator aims to produce more realistic results to deceive the discriminator in the opposite way. The generator (\(G_1, G_2\)) and the discriminator (\(D_I\)) use the same structure as the network proposed by Ganomaly [25] and define a new loss function.

The loss function proposed in this paper consists of three main components. The adversarial loss (\(L_{adv}\)) contributes to the overall training of the framework based on the GAN structure. The generator and the discriminator compete with each other, leading to better results. The discriminator classifies both real and fake images, with the fake output close to zero and the real output close to one. Binary cross-entropy, commonly used for binary classification, is applied to each output. Only the discriminator applies the loss function using Equation 1, which is based on binary cross-entropy. The generator defines the loss function to minimize the difference between each output based on L2 loss, as shown in Equation 3.

\[ L_{BCE}(a,b) = -\frac{1}{N} \sum_{i=0}^{n-1} [b \log(a) + (1-b) \log(1-a)] \quad (1) \]

\[ L_{adv}(D_I) = L_{BCE}(D_I(x), 1) + L_{BCE}(D_I(G_1(x)), 0) + L_{BCE}(D_I(G_2(x')), 0) \quad (2) \]

\[ L_{adv}(G) = \frac{1}{2} \mathbb{E}_{x \sim p_{data}(X)} [\|D_I(x) - D_I(G_1(x))\|_2 + \|D_I(x) - D_I(G_2(x'))\|_2] \quad (3) \]

Reconstruction loss (\(L_{rec}\)) defines a loss function to minimize the difference between the results created by the generator and the original image (Equation 4). The difference between all output images is constructed based on L1 loss. The minimization part with \(x\) induces the results of \(G_1(x)\) and \(G_2(x')\) to be similar to \(x\). Particularly, the difference from \(G_1(x)\) is very large, but it is essential because \(G_2(x')\) needs to produce some images for comparison. The minimization part between \(G_1(x)\) and \(G_2(x')\) listed at the end is the most related part to the score in the test, and it is characterized by having a similar difference in the reconstruction loss, thus being configured with the lowest ratio.

\[ L_{rec} = \frac{1}{3} \mathbb{E}_{x \sim p_{data}(X)} \left[\|x - G_1(x)\|_1 + \|x - G_2(x')\|_1 + \|G_1(x) - G_2(x')\|_1\right] \quad (4) \]

latent loss (\(L_{lat}\)) is defined with a similar meaning to the reconstruction loss. While reconstruction loss minimizes the differences between image domains, latent loss minimizes the differences between latent vectors in the feature space. The latent vector (\(z\)) is the output of the encoder (\(E_1, E_2, E_3\)) for each input image, and these vectors (\(z, z', z''\)) represent the compressed representation of the input image. The decoder plays a crucial role in restoring these vectors, as it aims to generate results that closely resemble the original input. Therefore, the latent loss helps each generator produce consistent results by minimizing the differences in the latent vectors. Each term is based on L2 loss and can be expressed as:

\[ L_{lat} = \frac{1}{3} \mathbb{E}_{z \sim p_{data}(Z)} [\|z - z'\|_2 + \|z - z''\|_2 + \|z' - z''\|_2] \quad (5) \]

Total loss including all of the above-mentioned adversarial loss, reconstruction loss, and Latent loss is expressed as Equation 6. Each loss function is multiplied by an appropriate hyperparameter to achieve the best performance. The values for each hyperparame-ter are mentioned in Section 4.

\[ L_{total} = \lambda_{adv} L_{adv} + \lambda_{rec} L_{rec} + \lambda_{lat} L_{lat} \quad (6) \]

\subsection{Test}
A process of determining whether the input image is normal or diseased is required after sufficient training has progressed. In the same way as the training framework, after outputting all results, the abnormal score is calculated using only two images generated by the generator, which is shown in Figure 3. 

\begin{figure}[htb!]
    \centering
    \includegraphics[width=15cm]{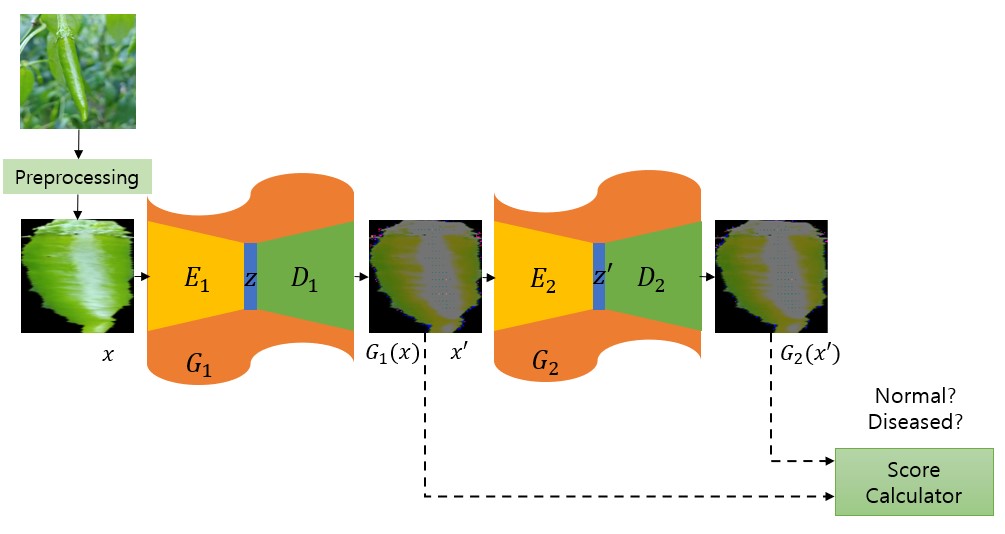}
    \caption{Test architecture using score calculator}
\end{figure}

The score calculator calculates the L2 loss for each RGB image, where \(G_1(x)\) and \(G_2(x')\) are three-dimensional color images generated by the generator. When the error \(e\) for each image is calculated, it is converted into a final score through a normalization process as shown in Equation 7. This score represents the closeness to zero, where a smaller error indicates a normal pepper image. Depending on the distribution of the entire experimental data, the threshold value \(\tau\) is chosen to maximize the false positive rate (FPR) and true positive rate (TPR) and determine whether the chili pepper image is normal or diseased.

\[ S(G_1(x), G_2(x')) = \frac{e - \min(e)}{\max(e) - \min(e)} \quad (7) \]

\section{Experiments}
\subsection{Datasets}
Among the data provided by AIHub[26], a dataset called "Outdoor Crop Disease Di-agnostic Image" is used. This dataset is diseased image data of 10 major open-field crops and provides JSON with coordinates for the location of the fruit or leaf (metadata). For the experiment, only the data corresponding to chili peppers are used among the various classes. However, in the constructed image data, a considerable number of data on chili leaves, not only chili fruits, are distributed. If the datasets are constructed using only the data on chili pepper berries, only 1,351 normal chili pepper images and 975 diseased chili pepper images remain. The network proposed in this paper is very insufficient because training should be conducted only with normal chili pepper images. Therefore, in this pa-per, we construct a train/test dataset by augmenting the remaining data. Data augmentation uses only rotation(degree of 90, 180, 270) and inversion to expand the number per data by 6-7 times for normal chili peppers. The finally constructed training data is 7,338 normal chili pepper images. And the test data consists of a total of 2,000 samples, 1,025 normal chili pepper images and 975 diseased chili pepper images.

\subsection{Detailed Training}
The experimental environment in this paper uses the RTX 3090 GPU in the operating system environment of Ubuntu 18.04 LTS. It uses the Tensorflow framework and conducts experiments based on version 2.6. In all experiments, Batchsize is fixed at 128, and Epoch is fixed at 2000. And the hyperparameters mentioned in \(L_{total}\), \(L_{rec}\) are set to 50 and set to 1 except \(L_{rec}\) to proceed with training. In addition, the length of the latent vector expressed by compressing the image is set to 100, and the initial learning rate is fixed at 0.0002. The generator and discriminator used in this paper proceed with the same structure as Ganomaly[25] and show higher performance based on this structure.

\subsection{Performance Evaluation}

We introduce the test results using the method proposed in this paper. First, the graph of the score for each data is shown in Figure 5. The graph in Figure 5 has a range of [0,1] and consists of 200 bins.

\begin{figure}[htb!]
    \centering
    \includegraphics[width=10cm]{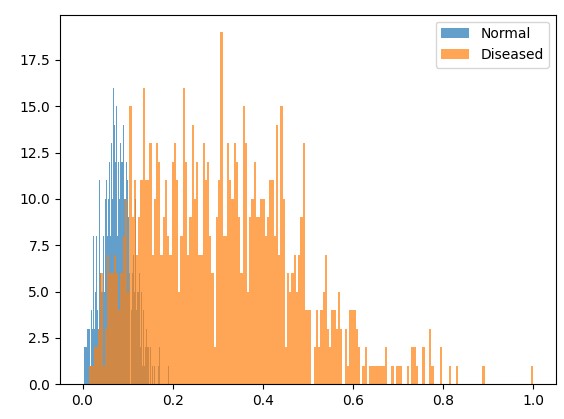}
    \caption{Score Distribution}
\end{figure}

The blue bar represents the score distribution of the normal pepper image, and the orange bar represents the score distribution of the diseased chili pepper image. Overall, each distribution shows a normal distribution with a convex middle region, and the maximum value of the normal chili pepper image does not exceed 0.2, showing a result close to 0. However, in the case of diseased pepper images, results close to 1 should be shown, but since they are evenly distributed in the range of [0, 1], there are overlapping distributions. Binary classification is performed using the optimal threshold \(\tau\) to distinguish between normal and diseased. The optimal threshold \(\tau\) is calculated using the difference between the true positive rate (TPR) and false positive rate (FPR), and the value calculated in this paper is 0.1402. As mentioned in Equation 8, if the score is less than \(\tau\), it is judged to be normal; otherwise, it is judged to be diseased.

Figure 6 shows the results of the proposed method, where column 1 shows the original image, and columns 2 and 3 show the results output by the generator. Lines 1 to 4 represent the results of the normal chili pepper image, while lines 5 to 8 represent the results of the diseased chili pepper image. The difference between the original image and the generated image is blurred or different in color, showing results that cannot be reconstructed in detail. It can be observed that the difference between \(G_1(x)\) and \(G_2(x')\) is not significant visually. Overall, the results show the approximate shape and color distribution through training, and particularly, areas where light is reflected tend to be enlarged and emphasized. 

The images in columns 4 to 6 represent the combined difference images using the results in columns 1 to 3. Column 4 shows \(|x - G_1(x)|\), column 5 shows \(|x - G_2(x')|\), and column 6 shows \(\alpha|G_1(x) - G_2(x')|\). \(|x - G_1(x)|\) represents the difference image between the original image and \(G_1(x)\), while \(|x - G_2(x')|\) represents the difference image between the original image and \(G_2(x')\). From the results, there is no significant difference, making it difficult to distinguish between lines 1 to 4 (normal) and lines 5 to 8 (diseased). Therefore, if this is used to calculate the score, it will yield low performance results. The results in column 6 (\(\alpha|G_1(x) - G_2(x')|\)) show the results of the weighted difference images between \(G_1(x)\) and \(G_2(x')\) used in this paper, where the difference is very small and the weight \(\alpha\) is set to 20. Looking at the results in column 6, it can be seen that the overall difference in \(\alpha|G_1(x) - G_2(x')|\) for the normal chili pepper image is very small, with the blue pixels slightly distributed. However, \(\alpha|G_1(x) - G_2(x')|\) for the diseased chili pepper image shows a noticeable difference, where not only blue pixels are distributed, but the green pixels appear intense in the diseased area. Therefore, it can be observed that the differential images of the generators can better divide the results between normal and diseased images.

Table 1 shows the method proposed in this paper and a performance comparison with previous studies. CAE has the same input dimension as the dataset, adding convolution to the autoencoder. It has a U-Net structure with skip-connections and defines a loss function for image reconstruction. Unlike Ganomaly[25] and the proposed method, the reconstruction image of CAE takes a different form from the original structure, resulting in lower performance with many misclassified results. The F1 score for CAE shows a low performance of 56.7\% (normal) and 68.6\% (diseased). Ganomaly[25] serves as a baseline model for the proposed method in this paper. It achieves high performance by calculating the difference between latent vectors as the score. However, when Ganomaly[25] is calculated based on image scores, it shows a low F1 score of 21.3\% for normal and 12.2\% for diseased. Although it clearly differs from the performance of CAE, it indicates the difficulty of classification based on images in the reconstruction method. However, the proposed method in this paper successfully addresses this challenge. It shows a higher F1 score compared to the CAE image reconstruction method, demonstrating a significant performance improvement of 23.7\% for normal and 13.7\% for diseased. While it outperforms the performance based on latent vectors, it is particularly noteworthy that it is achieved through an error method between images. The proposed method achieves a quantitative evaluation of 90.9\% for normal and 89.1\% for diseased in terms of F1 score. It also shows high precision and recall values, especially for normal and diseased images, with precision over 95\%. This indicates accurate classification of actual normal chili pepper images and very few misclassified results among the detected diseased images. Although there is a slight performance decline in precision for normal images and recall for diseased images, considering the overall F1 score, it can be regarded as superior performance.

\begin{table}[htbp]
 \caption{Comparisons with previous researches}
  \centering
  \begin{tabular}{lllll}
    \toprule
    Method     & Category     & Precision &Recall &F1-score \\
     \midrule
    CAE & Normal & 0.727 & 0.464 & 0.567 \\
        & Diseased & 0.592 & 0.816 & 0.686 \\
     \midrule
    Ganomaly[25] (based Feature) & Normal & 0.875 & 0.895 & 0.885 \\
                               & Diseased & 0.887 & 0.866 & 0.876 \\
     \midrule
    Ganomaly[25] (based Image) & Normal & 0.836 & 0.562 & 0.672 \\
                              & Diseased & 0.658 & \textbf{0.884} & 0.754 \\
     \midrule
    
    Proposed Method & Normal & 0.859 & \textbf{0.966} & \textbf{0.909} \\
                     & Diseased & \textbf{0.959} & 0.833 & \textbf{0.891} \\
    \bottomrule
  \end{tabular}
  \label{tab:table}
\end{table}

\begin{figure}[htb!]
    \centering
    \includegraphics[height=20cm]{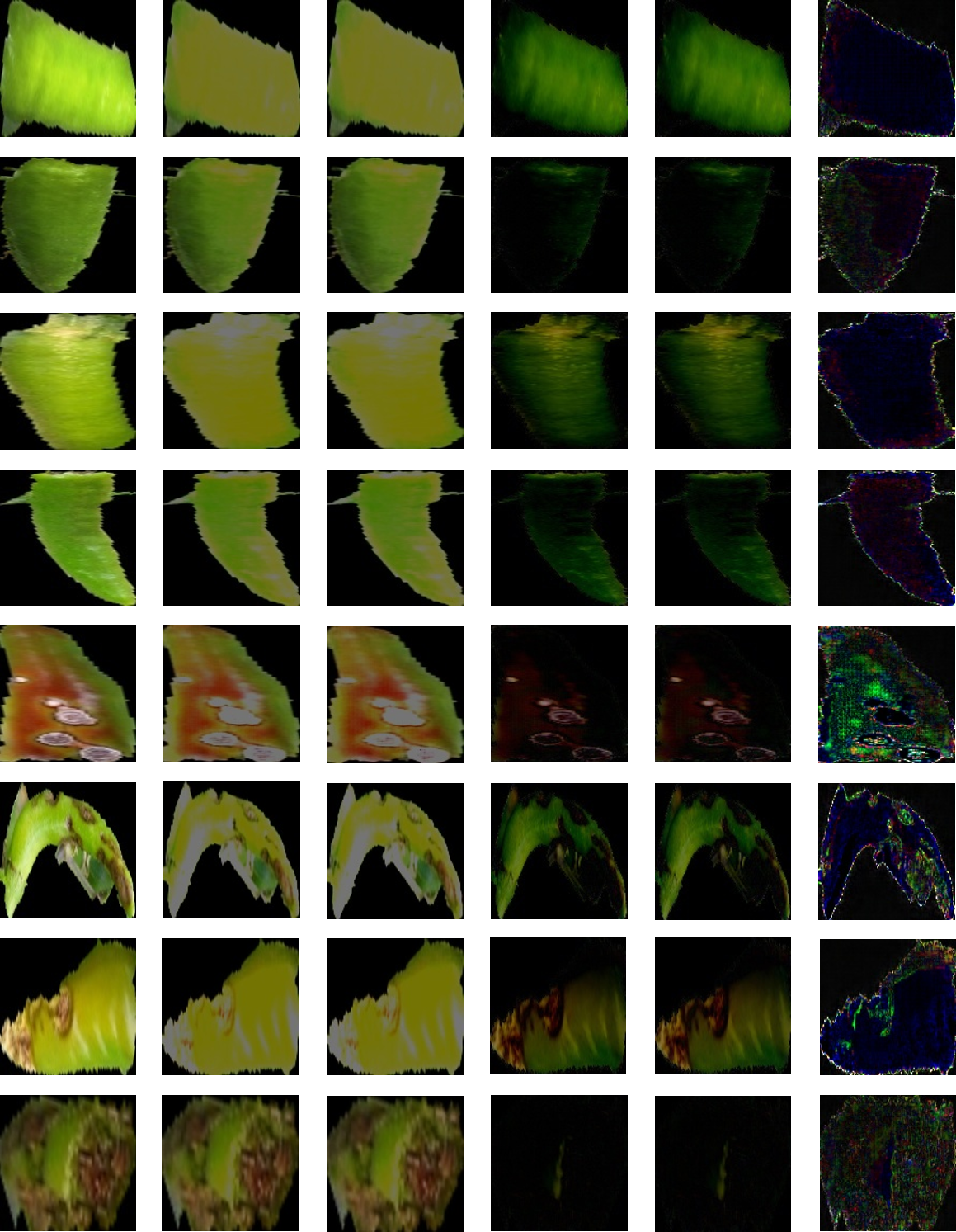}
    \caption{Reconstruction and subtraction results (Column 1: Input image, Column 2-3: Reconstruction image, Column 4-6: difference images}
\end{figure}

\subsection{Ablation Study}
In this paper, the final score was calculated using the results of $G_1$ and $G_2$. The diseased image was diagnosed using them. The form of a series in which two autoencoder structures were continuously combined and the corresponding score calculation method showed great performance improvement. This section also introduces the combined diagnostic results of each generator and the diagnostic results between the latent vectors. Latent vector is a part that is not noted in this paper, but the analysis is conducted to compare it with previous researches. The combination of the score calculation method is 3 latent vectors and 3 images, which are analyzed for a total of 6 types, and the evaluation method is compared using AUC, EER, AP, and Macro F1.

The AUC sets the True Positive Rate (TPR) and False Positive Rate (FPR) to $x$ and $y$ axes, respectively, drawing a Receiver Operating Characteristic (ROC) curve, and then obtains the area below. This area is a numerical value for performance evaluation. It means that the closer it is to 1, the better the model. Equal Error Rate (EER) is easily obtained using the ROC Curve and is one of the quick ways to compare accuracy. It means that TPR and FPR are the same ratio, and the lower EER, the more accurate the model is. Unlike the ROC Curve drawn with TPR and FPR, AP refers to the area under the graph drawn with Recall and Precision as $x$ and $y$ axes. AP is mainly used in the field of computer vision, and the higher the value, the better. Macro F1 refers to the average of F1 scores calculated by different positives among several classes. It means that the closer it is to 1, the higher performance.

\begin{figure}[htb!]
    \centering
    \includegraphics[width=15cm]{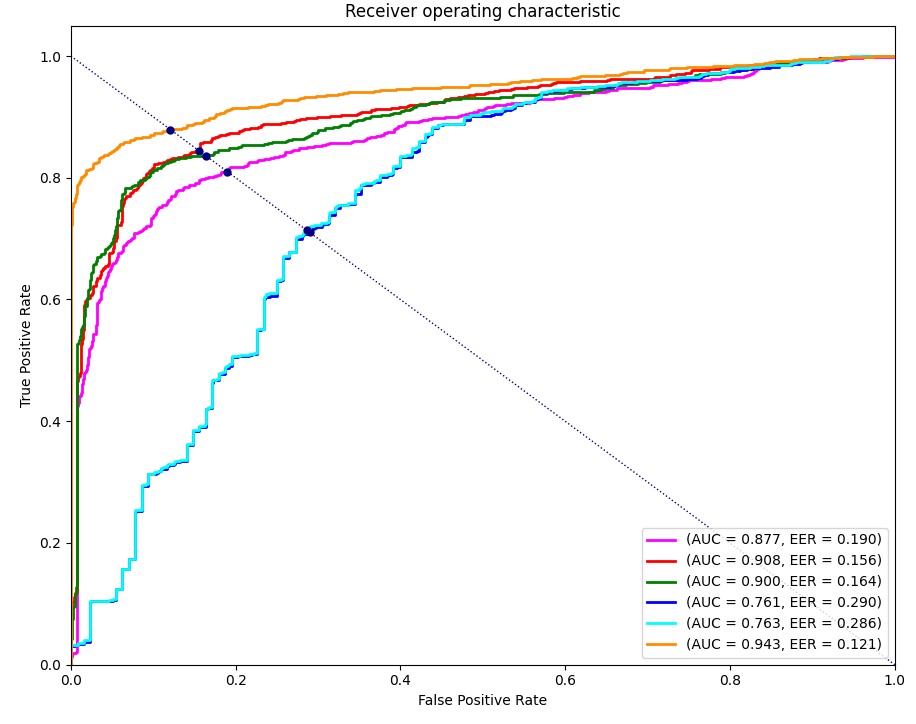}
    \caption{Comparison of ablation study with AUC and EER}
\end{figure}

Figure 7 shows the ROC Curve for the six score calculation methods. The Blue and Cyan graphs show results for $S(x,G_1(x))$ and $S(x,G_2(x'))$. The two results show a very slight difference on the graph. The AUC is 0.761 and 0.763, which is significantly lower than other score calculation methods and the EER is higher. Magenta, Red, and Green show a graph of the scoring method using a latent vector and show results that are significantly higher than scoring based on the image. Magenta is the graph of $S(z,z')$, Red is the graph of $S(z,z'')$, and Green is the graph of $S(z',z'')$, of which $S(z,z'')$ shows the best performance. The methods $S(G_1(x),G_2(x'))$ proposed in this paper are the highest performance methods with the highest AUC and lowest EER compared to other methods. AUC is 0.941 and EER is 0.121.

Table 2 shows additional comparison information between AP and Macro F1, including the above-mentioned AUC and EER. Like AUC and EER, $S(x,G_1(x))$ and $S(x,G_2(x'))$ show the lowest performance in AP and Macro F1. In the method of calculating the score based on the latent vector, $S(z,z'')$ performed the most with AP 0.915 and Macro F1 0.861. However, the method proposed in this paper shows the highest performance compared to all of these. AP is 0.959 and Macro F1 is 0.900, which is an increase of 0.44, 0.39 compared to $S(z,z'')$ which showed the best performance in the latent vector. Therefore, $S(G_1(x),G_2(x'))$ is compared with other score calculation methods and analyzed through several methods, indicating that it is the best method for all areas.

\begin{table}[htbp]
 \caption{Comparisons with previous researches}
  \centering
  \begin{tabular}{ccccccc}
    \toprule
    Method     & $S(z,z')$     & $S(z,z'')$ &$S(z',z'')$ &$S(x,G_1(x))$ &$S(x,G_2(x'))$ &$S(G_1(x),G_2(x'))$ \\
    \hline
    AUC & 0.877 & \underline{0.908} & 0.900 & 0.761 & 0.763 & \textbf{0.943} \\
    \hline
    EER & 0.190 & \underline{0.155} & 0.164 & 0.290 & 0.286 & \textbf{0.121} \\
    \hline
    AP & 0.879 & \underline{0.915} & 0.910 & 0.699 & 0.701 & \textbf{0.959} \\
    \hline
    Macro F1 & 0.826 & \underline{0.861} & 0.859 & 0.712 & 0.713 & \textbf{0.900} \\
    \hline
 
  \end{tabular}
  \label{tab:table}
\end{table}

\section{Conclusions}
\subsection{Datasets}
We show a GAN-based training structure in which two generators are connected in series and the discriminator distinguishes two fakes and one real. In addition, based on the chili pepper image, the background was removed using grabcut and a system for dis-ease diagnosis was shown in the image reconstruction method. The proposed method calculated and evaluated the score using the error between the reconstructed images rather than the latent vector. And it shows a very high performance improvement from the perspective of the image unlike previous researches. However, since the dataset used has a limited color or type of pepper, it is necessary to expand and conduct research. In addition, in order to automatically extract the bounding box coordinates required by grabcut, it will attract attention in the smart farm field if it is linked to automatically detecting the location of peppers in connection with object detection in the future.

\bibliographystyle{unsrt}

\end{document}